\documentclass[letterpaper, 10 pt, conference]{ieeeconf}  

\IEEEoverridecommandlockouts                              

\usepackage[T1]{fontenc}
\usepackage{times}
\usepackage{graphics} 
\usepackage{graphicx}
\usepackage{caption}
\usepackage{subcaption}
\usepackage{amsmath,amssymb,amsopn,amstext,amsfonts}
\usepackage{cancel}
\usepackage[space]{cite}
\usepackage{pdfsync}
\usepackage{balance}
\usepackage{color}
\usepackage{mathtools}
\usepackage{bm}

\usepackage{diagbox}
\usepackage{float}
\usepackage{epstopdf}
\usepackage{pifont}
\usepackage{fixltx2e}
\usepackage{amsmath}
\usepackage{multirow}
\usepackage{url}
\usepackage{verbatim}
\usepackage{caption}
\usepackage{adjustbox}
\usepackage{booktabs}
\usepackage{threeparttable}
\usepackage{makecell}
\usepackage{tabularx}
\usepackage{arydshln}
\usepackage[normalem]{ulem}
\usepackage{algorithm}
\usepackage{ifthen}

\usepackage{algorithmic}

\usepackage{array}
\usepackage{textcomp}
\usepackage{stfloats}

\usepackage[misc]{ifsym}

\usepackage{siunitx}
\pdfpxdimen=\dimexpr 1 in/72\relax  

\makeatletter
\let\NAT@parse\undefined
\makeatother
\usepackage[pagebackref=false,breaklinks=true,colorlinks,bookmarks=true,bookmarksnumbered=true]{hyperref}
\usepackage{cleveref}

\title{\LARGE \bf
Temporal and Rotational Calibration for Event-Centric \\Multi-Sensor Systems
}

\author{
Jiayao Mai$^{1}$*, Xiuyuan Lu$^{2}$*, Kuan Dai$^{1}$, Shaojie Shen$^{2}$ and Yi Zhou$^{1}$$^{\textrm{\Letter}}$
\thanks{$^{1}$School of Robotics, Hunan University, Changsha, China. Email: \{maijy, eeyzhou\}@hnu.edu.cn.}%
\thanks{$^{2}$Department of Electronic and Computer Engineering, Hong Kong University of Science and Technology, Hong Kong. Email: xluaj@connect.ust.hk.}%
\thanks{*Equal contribution; ${\textrm{\Letter}}$ Corresponding author: Yi Zhou.}
}

\DeclareMathOperator*{\argmax}{arg\,max}
\DeclareMathOperator*{\argmin}{arg\,min}

\global\long\def\bfb{\mathbf{b}}

\global\long\def\bfp{\mathbf{p}}

\global\long\def\bfx{\mathbf{x}}

\global\long\def\bfX{\mathbf{X}}

\global\long\def\cC{\mathcal{C}}

\global\long\def\cF{\mathcal{F}}

\global\long\def\cL{\mathcal{L}}

\global\long\def\cX{\mathcal{X}}

\global\long\def\Rot{\mathbf{R}}

\global\long\def\bfomega{\boldsymbol{\omega}}

\newcommand{\mai}[1]{\textcolor{black}{#1}}
\newcommand{\todo}[1]{\textcolor{black}{#1}} 

\newcommand{\Cov}{\mathrm{Cov}}
\newcommand{\Var}{\mathrm{Var}}

\begin{document}

\makeatletter
\makeatother
\maketitle
\begin{abstract}

Event cameras generate asynchronous signals in response to pixel-level brightness changes, offering a sensing paradigm with theoretically microsecond-scale latency that can significantly enhance the performance of multi-sensor systems. Extrinsic calibration is a critical prerequisite for effective sensor fusion; however, the configuration that involves event cameras remains an understudied topic. In this paper, we propose a motion-based temporal and rotational calibration framework tailored for event-centric multi-sensor systems, eliminating the need for dedicated calibration targets. Our method uses as input the rotational motion estimates obtained from event cameras and other heterogeneous sensors, respectively. Different from conventional approaches that rely on event-to-frame conversion, our method efficiently estimates angular velocity from normal-flow observations, which are derived from the spatio-temporal profile of event data. The overall calibration pipeline adopts a two-step approach: it first initializes the temporal offset and rotational extrinsics by exploiting kinematic correlations in the spirit of Canonical Correlation Analysis (CCA), and then refines both temporal and rotational parameters through a joint nonlinear optimization using a continuous-time parametrization in SO(3). Extensive evaluations on both publicly available and self-collected datasets validate that the proposed method achieves calibration accuracy comparable to target-based methods, while exhibiting superior stability over purely CCA-based methods, and highlighting its precision, robustness and flexibility. To facilitate future research, our implementation will be made open-source. Code: \url{https://github.com/NAIL-HNU/EvMultiCalib}.
\end{abstract}
\section{introduction}
Multi-sensor fusion is essential in robotics, enabling robust perception, localization, and decision-making by integrating complementary sensor data. 
IMUs provide high-frequency motion tracking but suffer from drift and lack spatial awareness. 
RGB cameras capture color and texture but struggle in low light and lack depth information. 
LiDAR offers precise depth sensing but is expensive and sensitive to weather conditions. 
In addition to these conventional sensors, event cameras provide a valuable complement with unique advantages. 
Inspired by biological vision, they detect asynchronous brightness changes, making them ideal for high-speed motion, extreme lighting, and HDR environments. 
Unlike frame-based cameras, they avoid motion blur, operate with low latency, and further enhance multi-sensor fusion in challenging robotic scenarios.


\begin{figure}[htbp]
    \centering
    \subcaptionbox{Temporal offset.}
    {\includegraphics[width=0.9\linewidth]{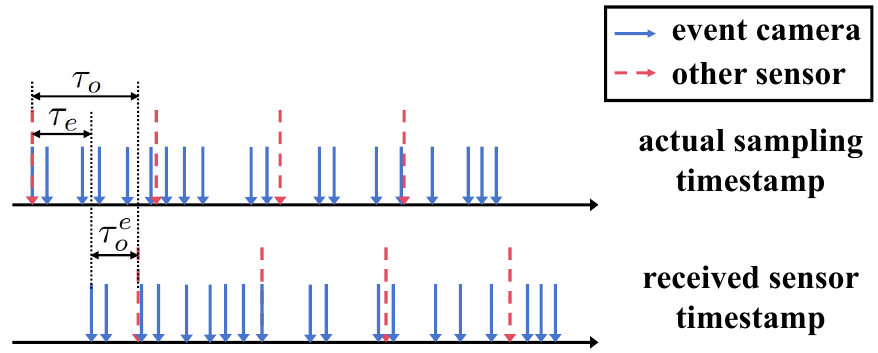}
    \label{fig:time_shift}}
    \hfill
    \vspace{5pt}
    \subcaptionbox{Spatial transformation.} 
    {\includegraphics[width=0.9\linewidth]{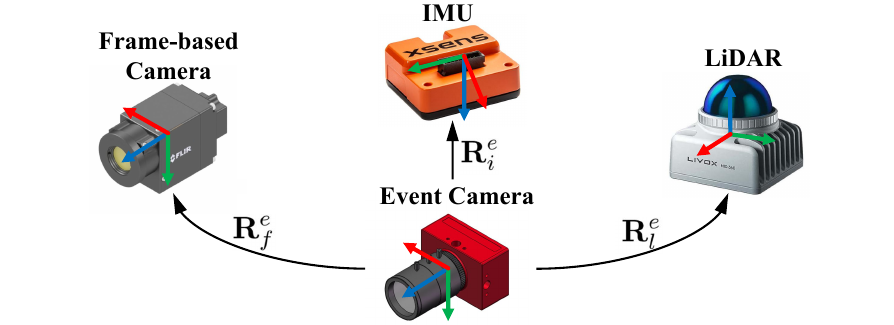}
    \label{fig:rotation}}
    \caption{Illustration of the extrinsic calibration task. (a) shows the temporal offset $\tau^e_o$ of the event camera with respect to the other sensors. \mai{$\tau_e$ and $\tau_o$ are the moments by which the sensors deviate from the actual sampling instants.} (b) demonstrates relative orientations ($\Rot^e_f$, $\Rot^e_i$, $\Rot^e_l$) of the event camera with respect to the frame-based camera, IMU, and LiDAR. }
    \label{fig:extrinsic_calib}
\end{figure}

To achieve effective multi-sensor fusion, extrinsic calibration is a prerequisite. 
As shown in Fig.~\ref{fig:extrinsic_calib}, extrinsic calibration typically consists of two key aspects: time offset estimation and spatial transformation estimation.
For time offset estimation, differences in trigger mechanisms, data transmission speeds, and sensor characteristics often cause misalignment between timestamps and actual sampling times. 
Temporal calibration aims to estimate and correct this offset, ensuring accurate synchronization across sensors.
For spatial transformation estimation, calibration is performed to align sensor data into a common reference frame, ensuring consistency and accuracy in the fusion system. 
While traditional calibration techniques are well-established for frame-based cameras and LiDARs, event cameras pose unique challenges due to their asynchronous, sparse, and intensity-independent nature. 
Existing approaches often preprocess event streams into image-like representations and then apply conventional calibration techniques designed for frame-based cameras. 
\todo{However, this transformation may introduce artifacts and degrade the event data’s temporal resolution.}
In this paper, we propose an event-centric extrinsic calibration framework leveraging the unique characteristics of event cameras which can directly operate on raw event streams, \todo{offering a targetless calibration approach.}
Our key contributions can be summarized as follows:
\begin{itemize}
    \item A novel motion-based extrinsic calibration framework for event-centric multi-sensor systems.
    By leveraging the latest advancements in geometric model fitting for event data, the proposed two-step method consists of a canonical correlation analysis (CCA)-based initialization method which exploits kinematic correlations across heterogeneous sensors, followed with a nonlinear optimization that jointly refines the temporal and rotational parameters in SO(3).
    \item Extensive evaluations on both public and self-collected datasets demonstrating the robustness and accuracy of our calibration method across various scenarios.
    \item An open-source multi-sensor extrinsic calibration toolbox that facilitates the calibration of event cameras with IMUs, frame-based cameras, and LiDARs.
\end{itemize}


\section{related work} \label{sec:related_work}
\begin{figure*}[!t]
\centering
\includegraphics[width=\linewidth]{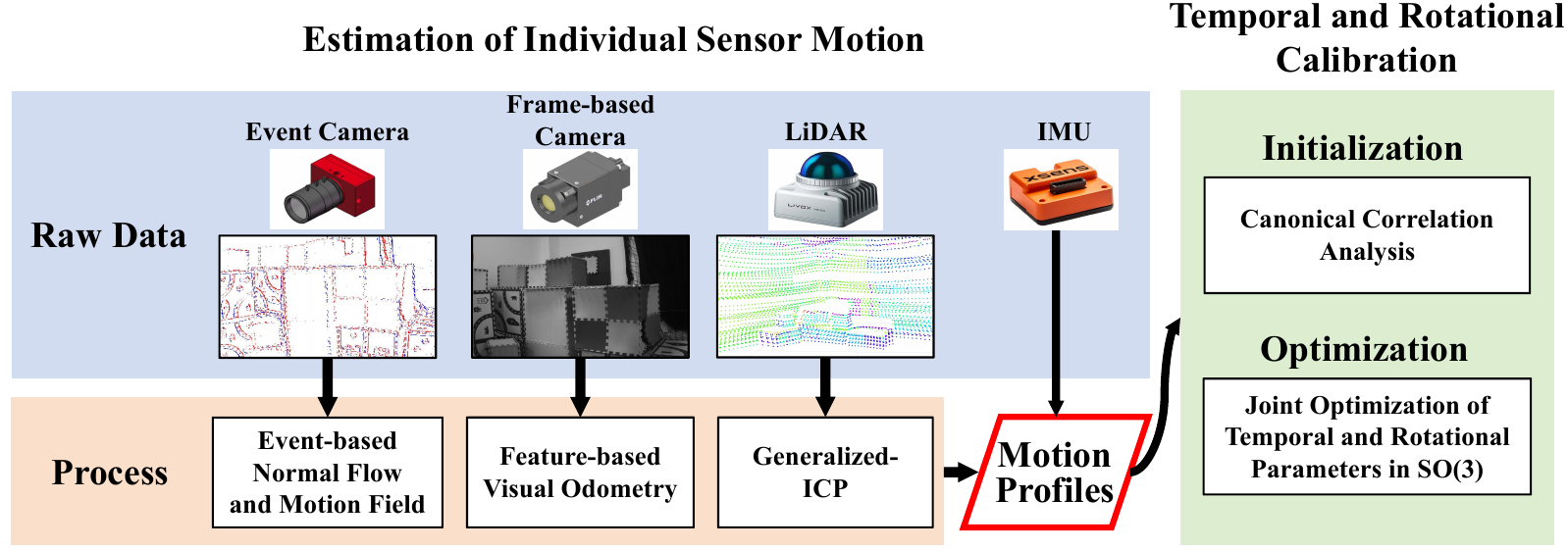}
\caption{\mai{Overview. Our calibration pipeline consists of two main stages. In the first stage, the motion features from each sensor are independently recovered. Following this, the temporal and rotational calibration is performed. Canonical Correlation Analysis (CCA) is used to initialize the extrinsic parameters, after which joint optimization of the extrinsic parameters is performed in SO(3) spline to ensure precise alignment and synchronization across the multi-sensor system.}}
\label{fig:overview}
\vspace{-2mm}
\end{figure*}
Several sensor calibration methods have been proposed in the past, among which the methods involving visual sensors can generally be classified into two categories: target-based methods and targetless methods.

\subsection{Target-Based Methods}
Target-based calibration methods rely on specific targets, such as checkerboard or AprilTag. Traditional visual methods can easily extract features directly from the pixel intensity in frames.
However, due to the unique trigger mechanism of event cameras, which can only capture dynamic information,~\cite{Mueggler2014iros, gossard2024icra, wang2024calib} use blinking LEDs or blinking screen to capture the calibration pattern. Similarly,~\cite{cai2024tim} utilizes a blinking screen to stimulate the event camera but employ frequency domain features to locate the calibration pattern. These methods require the event camera to remain relatively stationary with respect to the calibration pattern, and cannot calibrate sensors like the IMU that require motion excitation. 
In comparison,~\cite{Muglikar2021CVPR, jiao2023mech, huang2021iros, salah2024tip, wang2024ef} use event camera's ego-motion to trigger events. 
\todo{Among them,~\cite{Muglikar2021CVPR, jiao2023mech} reconstruct frames from events via E2VID~\cite{Rebecq19pami} and then use standard frame-based camera methods for calibration.}
\cite{huang2021iros, salah2024tip, wang2024ef}, based on prior knowledge of calibration patterns, fit pattern features through methods like clustering to perform calibration.

\subsection{Targetless Methods}

Targetless methods typically leverage the inherent correlation information in data from different sensors, such as motion, appearance (e.g., corner points and lines). This part
categorizes them into two types.

\subsubsection{Motion-based methods}
These methods first require the individual recovery of the pose or velocity of each sensor and then find the extrinsic parameters to align the recovered motion information.
\cite{qiu2021tro,li2024ral} use canonical correlation analysis to process the velocity between \todo{certain sensor pairs}, in which~\cite{li2024ral} transforms events to the image-like representation to recover the velocity. 
Hand-eye calibration~\cite{horaud1995ijrr} utilizes trajectory alignment to determine the extrinsic parameters, which can be extended to calibration involving event cameras.
\cite{taylor2016tro} also relies on the recovered poses and calculates the uncertainty of each pose, modeling the calibration problem as a maximum likelihood estimation problem. 
The challenge of applying the above method to event cameras lies in how to quickly and accurately recover the pose or velocity of event cameras.


\subsubsection{Appearance-based methods}
\todo{Some} methods calibrate by detecting representative features from the environment, such as edges and lines, and align them to perform calibration.
\cite{xing2023ral} extracts edges from event data and LiDAR point clouds and associates them to estimate the extrinsic parameters.
\cite{matao2021crlf} and~\cite{TUMTrafEvent2024tiv} are designed for intelligent transportation systems.
\cite{matao2021crlf} pairs line features from camera and LiDAR data, while~\cite{TUMTrafEvent2024tiv} pairs edges of moving objects in frame-based camera and event camera to recover the extrinsic rotation. 
\todo{Besides directly utilizing edge and line features in the environment, some methods define special metrics that can reflect appearance correlation.}
For instance,~\cite{censi2014icra} utilizes the cross-correlation between event rate and the magnitude of image intensity changes to calibrate the event-frame system. 
Similarly,~\cite{pandey2015jfr} proposes Mutual Information (MI) for calibrating vision and LiDAR, where the grayscale histogram and laser reflectivity serve as the probability distributions in MI. 
Building on this concept,~\cite{kevinta2023icra} extends MI to event cameras, but it requires sensors to remain stationary, with events triggered only by the laser emitted by the LiDAR.
Additionally,~\cite{taylor2015jfr} calibrates the frame-frame or frame-LiDAR system using a gradient orientation measure. In contrast,~\cite{cocheteux2024CVPR} directly employs a learning-based method to acquire the implicit correlation between raw event data and LiDAR point clouds. 
\todo{However, such methods generally require sensors to have overlapping fields of view and are typically limited by the need for well-structured environments and suitable feature correspondences.}

Building on previous work, this paper presents a motion-based approach. 
Compared with~\cite{Muglikar2021CVPR, jiao2023mech, huang2021iros, salah2024tip, wang2024ef}, which rely on predefined patterns, our method directly extracts the ego-motion from the raw event stream, offering greater flexibility. 
Unlike~\cite{qiu2021tro,li2024ral}, we recover the motion of the event camera by leveraging the differential and asynchronous features of the raw event stream, and ultimately use a nonlinear optimization method to provide more stable calibration results.


\section{problem statement} \label{sec:problem_statement}
\mai{This section addresses the problem of extrinsic calibration in a multi-sensor system.}
\todo{Given the intrinsic parameters of each individual sensor as prerequisites, our goal is to propose the extrinsic calibration in an event-centric multi-sensor suite, which may include the IMU, frame-based camera, and LiDAR.} 
As shown in Fig.~\ref{fig:overview}, our extrinsic calibration pipeline can be divided into the following two subproblems.

\subsection{Estimation of Individual Sensor Motion}
The first step is to recover the motion state of the individual sensor. 
Assuming the multi-sensor platform undergoes pure rotational motion, and building on the first principles of the event camera, the angular velocity is taken as the event camera's first-order kinematic state. 
Since the IMU directly outputs angular velocity, it is considered the first-order kinematic state of the IMU as well. 
For sensors with synchronized sampling at fixed time intervals, such as frame-based cameras, the relative rotation is regarded as their kinematic states. 
Each sensor's data is processed individually until the motion information accumulated is adequate for analyzing motion correlation.


\subsection{Temporal and \todo{Rotational} Calibration}
At this stage, we perform calibration under the following assumptions: all sensors are installed on a rigidly connected platform, and time offset is assumed to remain constant throughout the observation period. 
\todo{The motion-based extrinsic parameter calibration is divided into two steps.
In the initialization step, canonical correlation analysis is employed to provide an initial estimate of the time offset and extrinsic rotation for each individual sensor pair, ensuring quick convergence.
In the optimization step, we employ a continuous-time representation and utilize B-splines to model the rotational motion, and calibrate time offset and extrinsic rotation for all sensors simultaneously through a nonlinear optimization process.}

\section{Estimation of  Sensor Motion} \label{sec:sensor_ego_motion}

\subsection{Event-based Camera Angular Velocity Estimation}
To exploit the differential nature of event cameras, we estimate the angular velocity using event-based normal flow and the motion field equation~\cite{LU-RSS-24,ren2024eccv}. 
Assuming that the event data $\mathcal{E}\doteq\{(x_i,y_i,t_i)\}_{i=1}^N$ in the spatio-temporal domain near coordinate $\mathbf{x}_k=(x_k,y_k)$ at time $t_k$ on the time surface can be fitted to a local plane $\mathcal{T}:ax + by + c = t $, the normal flow $\mathbf{n}$ at coordinate $\mathbf{x}_k$ is calculated as

\begin{equation}\label{nflow}
\mathbf{n}=\frac{\nabla_\mathbf{x}\mathcal{T}}{\Vert\nabla_\mathbf{x}\mathcal{T}\Vert^2},
\end{equation}
where $\nabla_\mathbf{x}\mathcal{T}$ is the spatial derivative $[\frac{\partial\mathcal{T}}{\partial x},\frac{\partial\mathcal{T}}{\partial y}]^T$.
We calculate the covariance of the coefficients of the local plane $\mathcal{T}$ using the linear regression model: 

\begin{equation}\label{plane_cov}
\Cov(\mathbf{p}) = \frac{\sigma^2(\mathbf{X}^T\mathbf{X})^{-1}}{N-3},
\end{equation}
where $\bfp = \begin{bmatrix} a & b & c  \end{bmatrix}^T$, $\bfX=\begin{bmatrix} x_1 & y_1 & 1 \\ x_2 & y_2 & 1 \\ \vdots & \vdots & \vdots \\ x_N & y_N & 1\end{bmatrix}$, and $\sigma^2=\sum_{i=1}^{N}{(t_i-c-ax_i-by_i)}^2$. 
Therefore, we can propagate the covariance of plane to the variance of normal flow norm~($\Vert\mathbf{n}\Vert=\frac{1}{\sqrt{a^2+b^2}}$) by linearizing the norm:

\begin{equation}\label{nflow_cov}
\Var(\Vert\mathbf{n}\Vert)\approx\frac{\partial\Vert\mathbf{n}\Vert}{\partial\mathbf{p}}\Cov(\mathbf{p})(\frac{\partial\Vert\mathbf{n}\Vert}{\partial\mathbf{p}})^T.
\end{equation}
We remove data with large variance as outliers, and in our implementation, 20\% of the normal flow is discarded.

Then we can use the normal flow to calculate the angular velocity $\mathbf{\omega}$ according to the motion field equation~\cite{Trucco98book}:
\begin{equation}\label{motion_field}
\mathbf{u}(\mathbf{x})=\frac{1}{Z(\mathbf{x})}\mathbf{A}(\mathbf{x})\upsilon+\mathbf{B}(\mathbf{x})\mathbf{\omega}.
\end{equation}
Since the event camera undergoes pure rotational motion, and optical flow can be decomposed into normal flow and a component perpendicular to the normal flow, Eq.~\eqref{motion_field} can be rewritten as

\begin{equation}\label{ev_vel}
\Vert\mathbf{n}(\mathbf{x})\Vert^2=\mathbf{n}(\mathbf{x})^T\mathbf{B}(\mathbf{x})\mathbf{\omega}.
\end{equation}
Given a set of normal flows as input, Eq.~\eqref{ev_vel} can calculate the angular velocity of the event camera. For robust estimation, we employ RANSAC~\cite{Fischler81cacm} in this process.

\subsection{Frame-based Camera Angular Velocity Estimation}
We use a standard feature-based visual odometry approach to recover the \todo{relative} rotation.
As a first step, we extract ORB features~\cite{rublee2011orb} from the image and use the Hamming distance as the metric for brute-force matching. 
In this implementation, we ignore point pairs whose Hamming distance is larger than twice the minimum distance, \todo{which allows us to eliminate the majority of mismatches while retaining a sufficient number of matched point pairs.} 
In the second step, we recover the rotation by minimizing the reprojection error. To reduce the impact of mismatches, RANSAC~\cite{Fischler81cacm} is adopted \todo{to iteratively estimate the relative rotation.}


\subsection{Lidar Angular Velocity Estimation}
In order to determine the rotation from one scan to the next, we use the Generalized-ICP (GICP)~\cite{segal2009generalized} algorithm, \todo{an improved version of iterative closest point (ICP) that integrates point-to-point ICP and point-to-plane ICP, while introducing a probabilistic model to enhance the robustness and performance of the algorithm.}
To begin with, to correct for motion distortion, we preprocess the LiDAR measurements by projecting each point sampled in a scan onto the scan-end of this LiDAR frame using the previously estimated motion. 
Subsequently, we use the open-source small-GICP library~\cite{small_gicp} to downsample the point clouds and then match the source and target point clouds to recover the LiDAR relative rotation. 
\section{temporal and \todo{Rotational} Calibration} 
\label{sec:calibration}

Section~\ref{sec:sensor_ego_motion} discusses how to recover the motion state of the individual sensor. 
This section introduces how motion correlation is applied for extrinsic calibration. 
\mai{To begin, we apply correlation analysis to generate an initial estimate for extrinsic parameters. Subsequently, we refine these estimates through joint optimization in SO(3) spline.}

\subsection{Initialization}
 Canonical correlation analysis (CCA) and trace correlation are used to measure the three-dimensional correlation between $\omega_e$ and $\omega_o$, the two sets of angular velocities of the event camera and any other sensor, which allows us to obtain initial estimates of the temporal offset and extrinsic rotation by maximizing the correlation coefficient~\cite{qiu2021tro}.

In our implementation, for the sensors that only recover the relative rotation, the average angular velocity is obtained by dividing the relative rotation vector by the time interval, which is then used as the angular velocity at the intermediate moment between the two frames. 
Furthermore, cubic curve interpolation is used to convert the two sets of angular velocities to the same frequency. 
Let $\{\bfomega_e(t_i)\}^{N-1}_{i=0}$ and $\{\bfomega_o(t_i)\}^{N-1}_{i=0}$ represent $N$ angular velocities of the corresponding sensors at the sampling times $t_i$. $\tau^e_o$ denotes the time offset, and $\Rot^e_o$ denotes the extrinsic rotation of the event frame with respect to the other sensor frame.
The cross-covariance and auto-covariance can be calculated by


\begin{equation}\label{cov}
\begin{aligned}
\Sigma_{\bfomega_e\bfomega_e}&\approx\frac{1}{N-1}\sum^{N-1}_{i=0}(\bfomega_e(t_i)-\bar{\bfomega}_e)(\bfomega_e(t_i)-\bar{\bfomega}_e)^T, \\
\Sigma_{\bfomega_o\bfomega_o}&\approx\frac{1}{N-1}\sum^{N-1}_{i=0}(\bfomega_o^\prime(t_i)-\bar{\bfomega^\prime_o})(\bfomega_o^\prime(t_i)-\bar{\bfomega^\prime_o})^T, \\
\Sigma_{\bfomega_e\bfomega_o}&\approx\frac{1}{N-1}\sum^{N-1}_{i=0}(\bfomega_e(t_i)-\bar{\bfomega}_e)(\bfomega_o^\prime(t_i)-\bar{\bfomega^\prime_o})^T, \\
\Sigma_{\bfomega_o\bfomega_e}&\approx\frac{1}{N-1}\sum^{N-1}_{i=0}(\bfomega_o^\prime(t_i)-\bar{\bfomega^\prime_o})(\bfomega_e(t_i)-\bar{\bfomega}_e)^T,
\end{aligned}
\end{equation}
where $(\cdot)^\prime$ is the time offset compensated sample, $\bar{(\cdot)}$ is the mean of samples. 
Then the time offset can be calculated by 
\begin{equation}\label{CCA_time_offset}
{\tau^e_o}^*=\argmax_{\tau^e_o} \bar{r}(\bfomega_e, \bfomega_o),
\end{equation}
where $\bar{r}(\bfomega_e, \bfomega_o)$ is the trace correlation, and according to Eq. \eqref{cov}, it can be written as~\cite{qiu2021tro}
\begin{equation}\label{trace_corr}
\bar{r}(\bfomega_e, \bfomega_o)=\sqrt{(\frac{1}{3}Tr(\Sigma_{\bfomega_e\bfomega_e}^{-1}\Sigma_{\bfomega_e\bfomega_o}\Sigma_{\bfomega_o\bfomega_o}^{-1}\Sigma_{\bfomega_o\bfomega_e}))},
\end{equation}
where $Tr(\cdot)$ denotes the trace of matrix. In our implementation, we solve Eq. \eqref{CCA_time_offset} by enumerating time offsets within a specified range, with a step size of 10 ms. 
After compensating for the time offset between two sensors, the corresponding extrinsic rotation $\Rot^e_o$ can be calculated under the CCA framework: 
\begin{equation}\label{CCA_R}
\Rot^e_o = U \begin{pmatrix}
            1 & 0 & 0  \\
            0 & 1 & 0 \\
            0 & 0 & \det(UV^T) \end{pmatrix} V^T,
\end{equation}
where $U$ and $V$ is the left and right unitary matrix of the singular value decomposition of matrix $\Sigma_{\bfomega_o\bfomega_o}^{-1}\Sigma_{\bfomega_o\bfomega_e}$.

\subsection{Optimization}
To handle asynchronous data and facilitate time offset estimation, we adopt a continuous-time representation and use B-splines to model rotational motion in the optimization.
We employ a uniform cumulative cubic B-spline~\cite{Mueggler18tro} to represent the event camera's rotational trajectory as:
\begin{equation}
\begin{aligned}
    \Rot(t) &= \Rot_{0} \prod_{i=1}^{3} \exp \left( \tilde{\mathbf{B}}_j(u(t)) \mathtt{\Omega}_{i-1} \right), \\
    \addlinespace[10pt]
    \mathtt{\Omega}_i &= \Rot^e_f \left( \Rot_{i-1}^{\top} \Rot_i \right), \\
    \addlinespace[10pt]
    \tilde{\mathbf{B}}(u(t)) &= \frac{1}{6}
    \begin{pmatrix}
        6 & 0 & 0 & 0 \\
        5 & 3 & -3 & 1 \\
        1 & 3 & 3 & -2 \\
        0 & 0 & 0 & 1
    \end{pmatrix}\!
    \begin{pmatrix}
        1 \\
        u(t) \\
        u(t)^2 \\
        u(t)^3
    \end{pmatrix},
\label{eq:cubic_interp}
\end{aligned}
\end{equation}
where $\Rot_{i} \in \mathit{SO}(3)$ denotes the corresponding rotation control pose and $\tilde{\mathbf{B}}_j$ the $j$-th entry (0-based) of cumulative basis function.
$u(t)$ is the time normalization function~\cite{deBoor78book} that maps the time $t$ to the spline's parameter domain through basis translation.
We make use of the efficient method proposed in~\cite{Sommer2020cvpr} to compute the angular velocity $\dot\Rot(t)$ and related analytical derivatives.

Our goal is to obtain a B-spline curve that closely aligns with the motion estimation of all sensors by accurately estimating the extrinsic rotation and time offsets.
The full state vector in the optimization is defined as: 
\begin{equation}
\label{eq: state vector}
\begin{aligned}
    \bf{\cX} &= [\Rot_0, \hspace{1mm} \Rot_1, \hspace{1mm} \cdots, \hspace{1mm} \Rot_n, \hspace{1mm} \bfx_u, \hspace{1mm} \bfx_{imu} ], \\
    \addlinespace[5pt]
    \bfx_u &= [\Rot^e_f, \hspace{1mm} \tau^e_f, \hspace{1mm} \Rot^e_l, \hspace{1mm} \tau^e_l], \\
    \addlinespace[5pt]
    \bfx_{imu} &= [\Rot^e_i, \hspace{1mm} \tau^e_i, \hspace{1mm} \bfb_{\omega}],
\end{aligned}
\end{equation}
where $\bfx_u$ denotes the extrinsic rotations and time offsets between the event camera and the standard frames or LiDAR, respectively.
$\bfx_{imu}$, in addition to the extrinsic rotation and time offset between the event camera and IMU, includes an additional gyroscope bias term $\bfb_{\omega}$, which is assumed to remain constant throughout the data sequences.
n is the number of control poses used to fit the motion of the entire data sequence.

Given the angular velocity estimation from multiple sensors, the final objective function is defined as follows,
\begin{equation}
\label{eq: objective function}
\begin{aligned}
\bf{\cX}^{\star} &= \argmin_{\bf{\cX}} \cC,
\end{aligned}
\end{equation}
where the cost is defined as
\begin{equation}
\label{eq:new energy function}
\begin{aligned}
\cC &\doteq \sum_{m}\Vert \dot \Rot (t_m) - \bfomega_e(t_m) \Vert^2 \\
&+ \sum_{n}\Vert \dot  \Rot (t_n + \tau^e_n ) - \Rot^e_i\left(\bfomega_i(t_n) + \bfb_{\omega}\right) \Vert^2 \\
&+ \sum_{(i,j) \in \cF}\Vert \log\left( {\Rot^e_f}^T {\Rot (t_i + \tau^e_f )}^T \Rot (t_j + \tau^e_f )  \Rot^e_f \Rot^{c_j}_{c_i}\right) \Vert^2 \\
&+ \sum_{(p,q) \in \cL}\Vert \log\left( {\Rot^e_l}^T {\Rot (t_p + \tau^e_l )}^T \Rot (t_q + \tau^e_l )  \Rot^e_l \Rot^{l_q}_{l_p}\right) \Vert^2. \\
\end{aligned}
\end{equation}
$\bfomega_e(t_m)$ and $\bfomega_i(t_n)$ represent the estimated angular velocity of the event camera and the IMU angular velocity measurements, respectively.
$\cF$ and $\cL$ denote the frame pairs and LiDAR scan pairs with estimated pose changes.
\todo{
The first term represents the error between the angular velocity derived from the camera's rotational trajectory and the angular velocity estimated by the event camera. 
The second term reflects the error between the angular velocity derived from the camera's rotational trajectory and the angular velocity readings from the IMU. 
The third term captures the difference between the rotational change in the camera's trajectory and the rotation change obtained from the RGB camera.
The fourth term represents the difference between the rotational change in the camera's trajectory and the rotation change estimated by the LiDAR.
}
Ceres Solver~\cite{ceres-solver} is used to solve the non-linear optimization problem.
Our method makes no assumptions about the scene and can be easily extended to other sensors capable of angular velocity or rotation estimation for time offset and rotational extrinsic calibration.
\begin{figure}[t]
\centering
\includegraphics[width=0.45\textwidth]{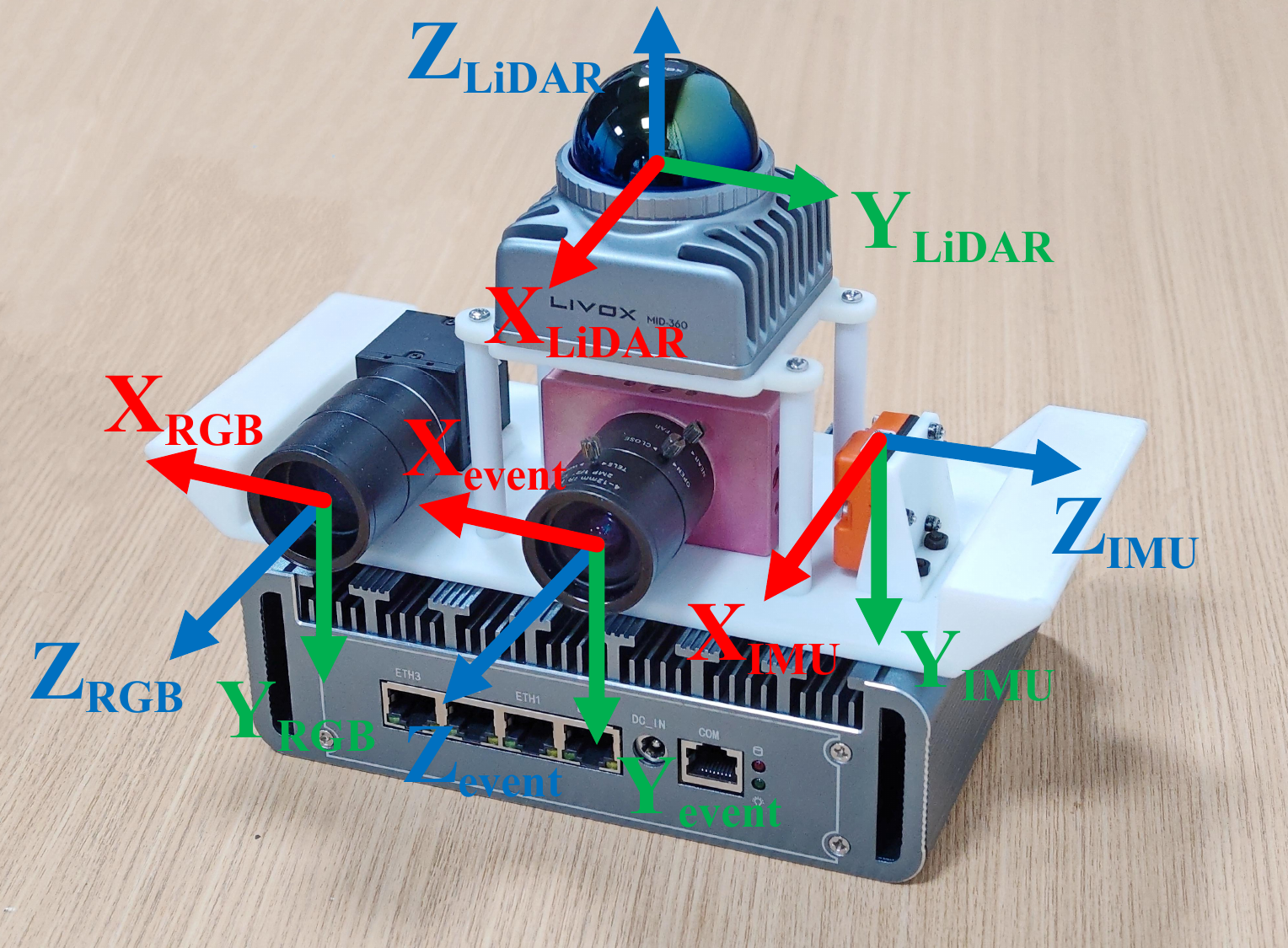}
\caption{\mai{A self-assembled multi-sensor platform integrating an event camera, an RGB camera, a LiDAR, and an IMU for comprehensive data acquisition. The coordinate systems of each sensor are explicitly labeled in the image, highlighting their spatial orientation within the platform.}}
\label{fig:sensor_setup}
\end{figure}
\begin{figure}[t]
  \centering
  \begin{subfigure}[t]{0.32\linewidth}
    \setlength{\abovecaptionskip}{2pt}
    \includegraphics[width=\linewidth]{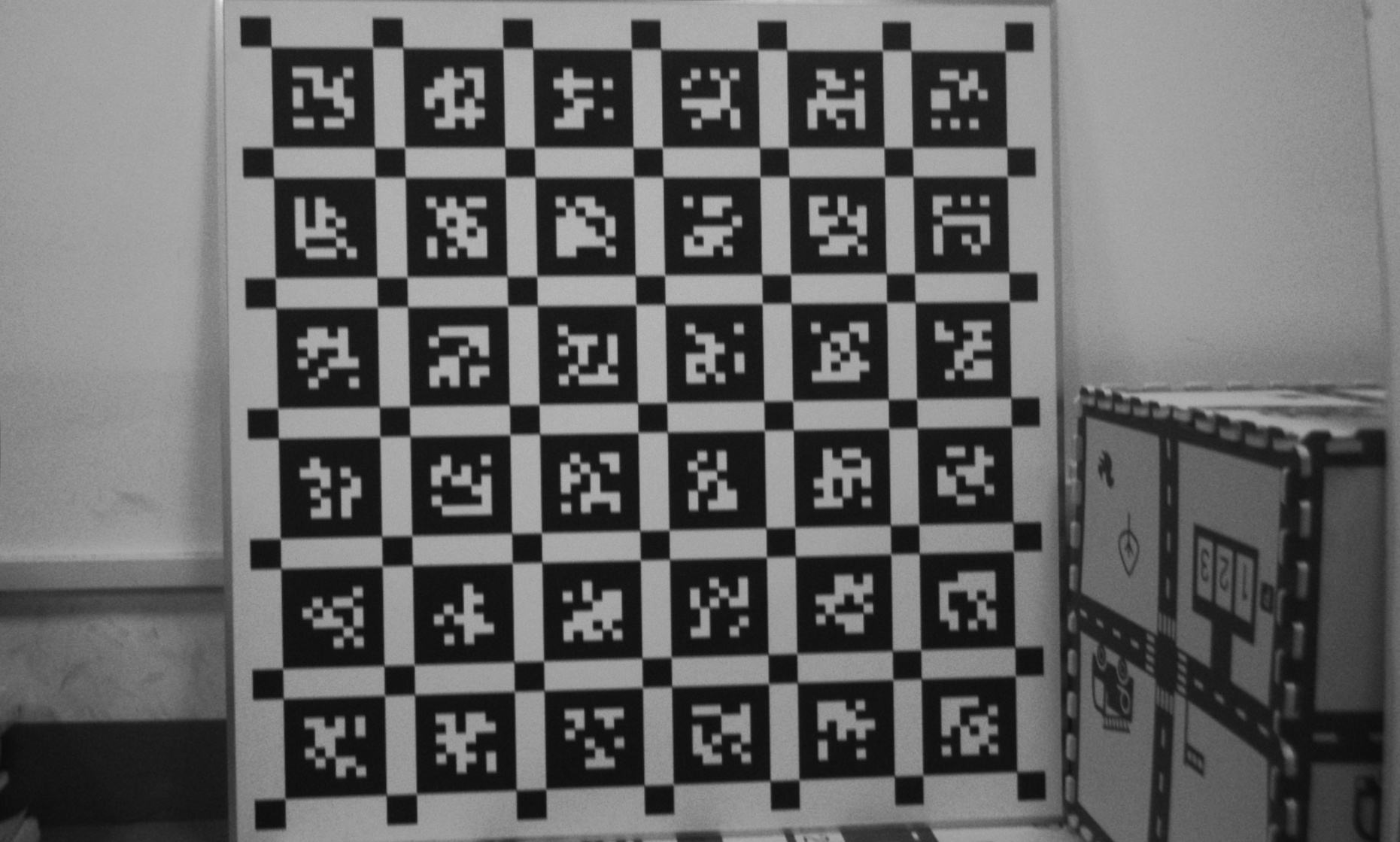}
    \caption{AprilTag}
    \label{fig:AprilTag}
  \end{subfigure}
  \begin{subfigure}[t]{0.32\linewidth}
    \setlength{\abovecaptionskip}{2pt}
    \includegraphics[width=\linewidth]{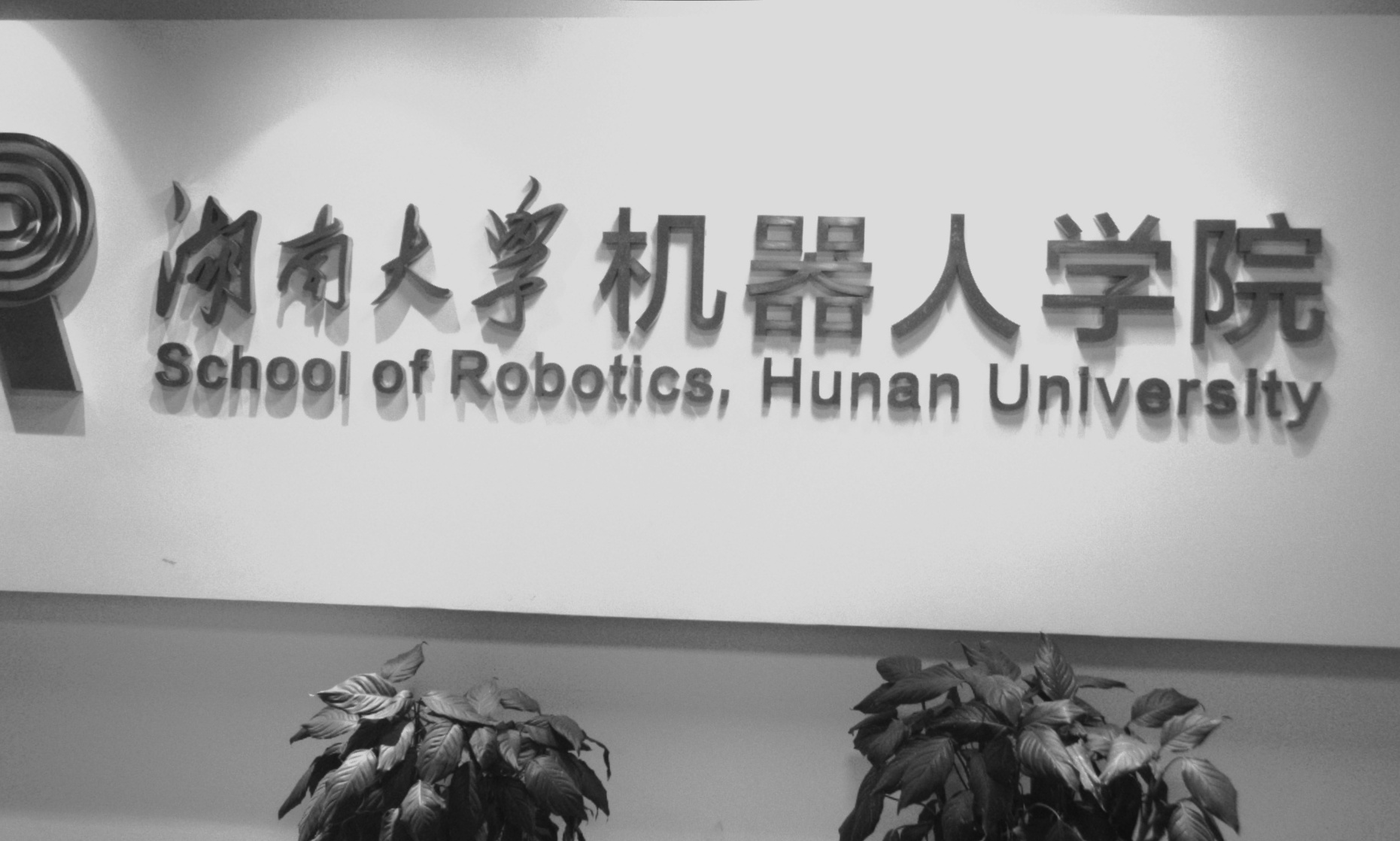}
    \caption{logo}
    \label{fig:logo}
  \end{subfigure}
  \begin{subfigure}[t]{0.32\linewidth}
        \setlength{\abovecaptionskip}{2pt}
    \includegraphics[width=\linewidth]{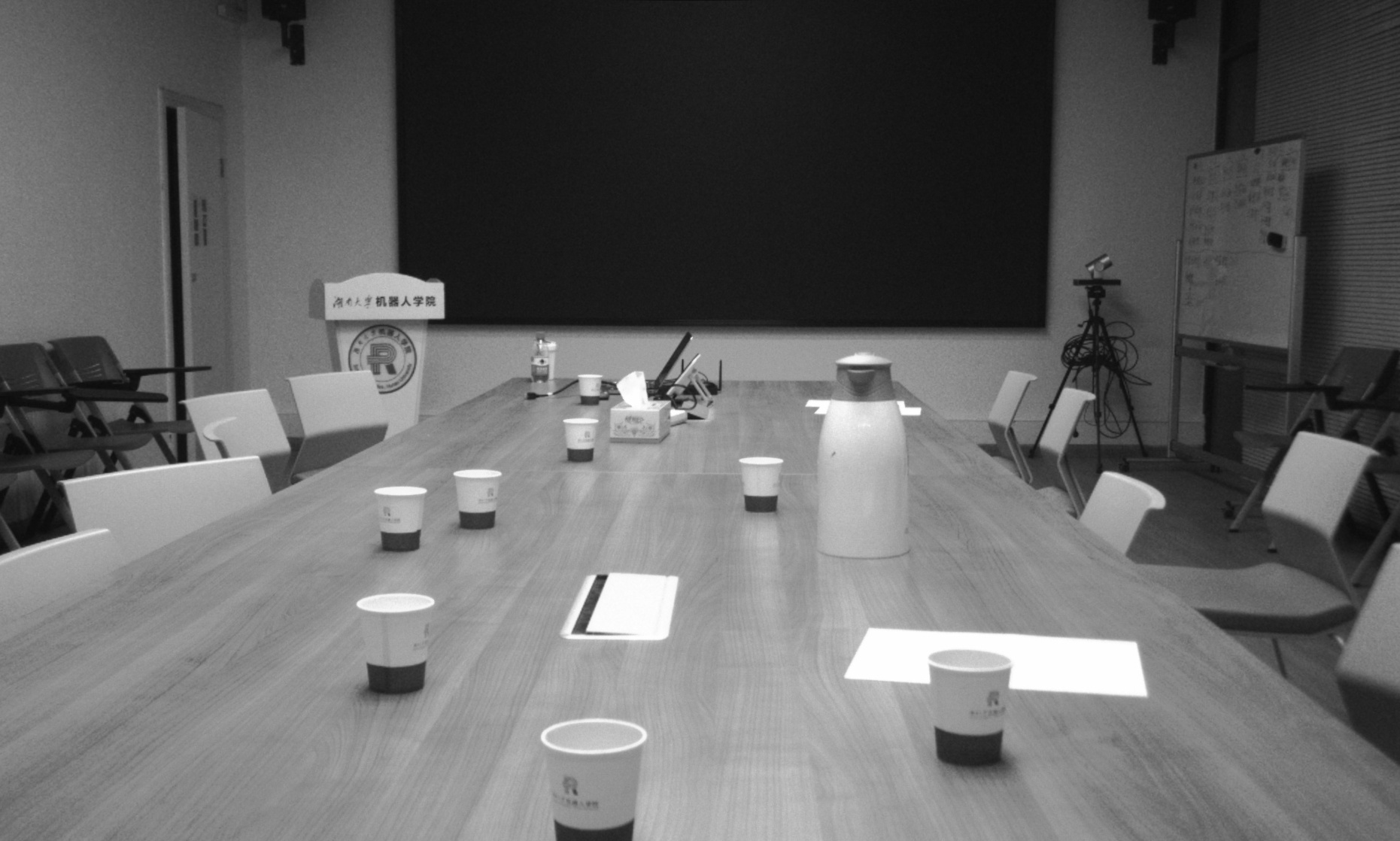}
    \caption{boardroom}
    \label{fig:boardroom}
  \end{subfigure}
  \hfill
  \vspace{1pt}
  \begin{subfigure}[t]{0.32\linewidth}
      \setlength{\abovecaptionskip}{2pt}
    \includegraphics[width=\linewidth]{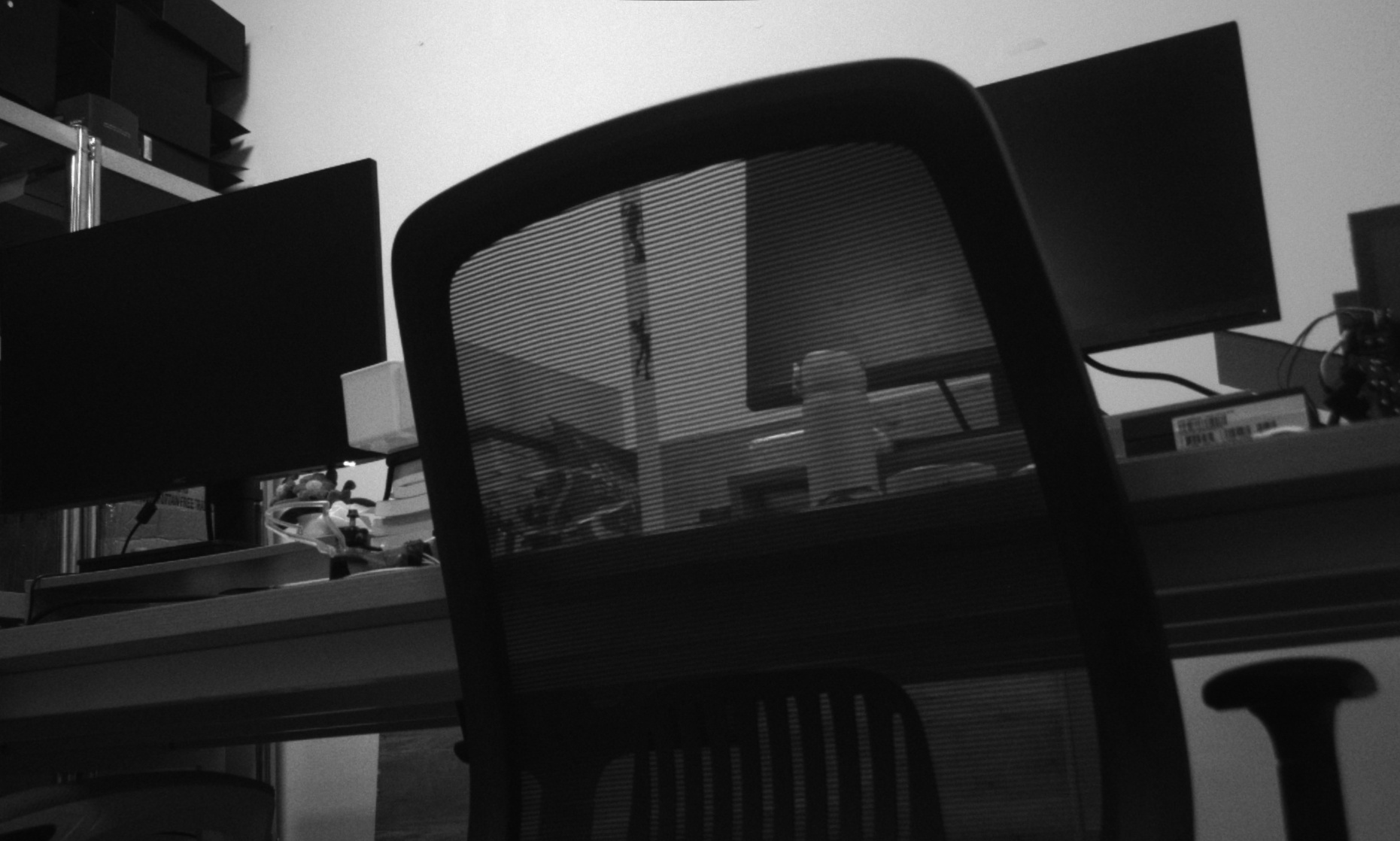}
    \caption{office}
    \label{fig:office}
  \end{subfigure}
  \begin{subfigure}[t]{0.32\linewidth}
      \setlength{\abovecaptionskip}{2pt}
    \includegraphics[width=\linewidth]{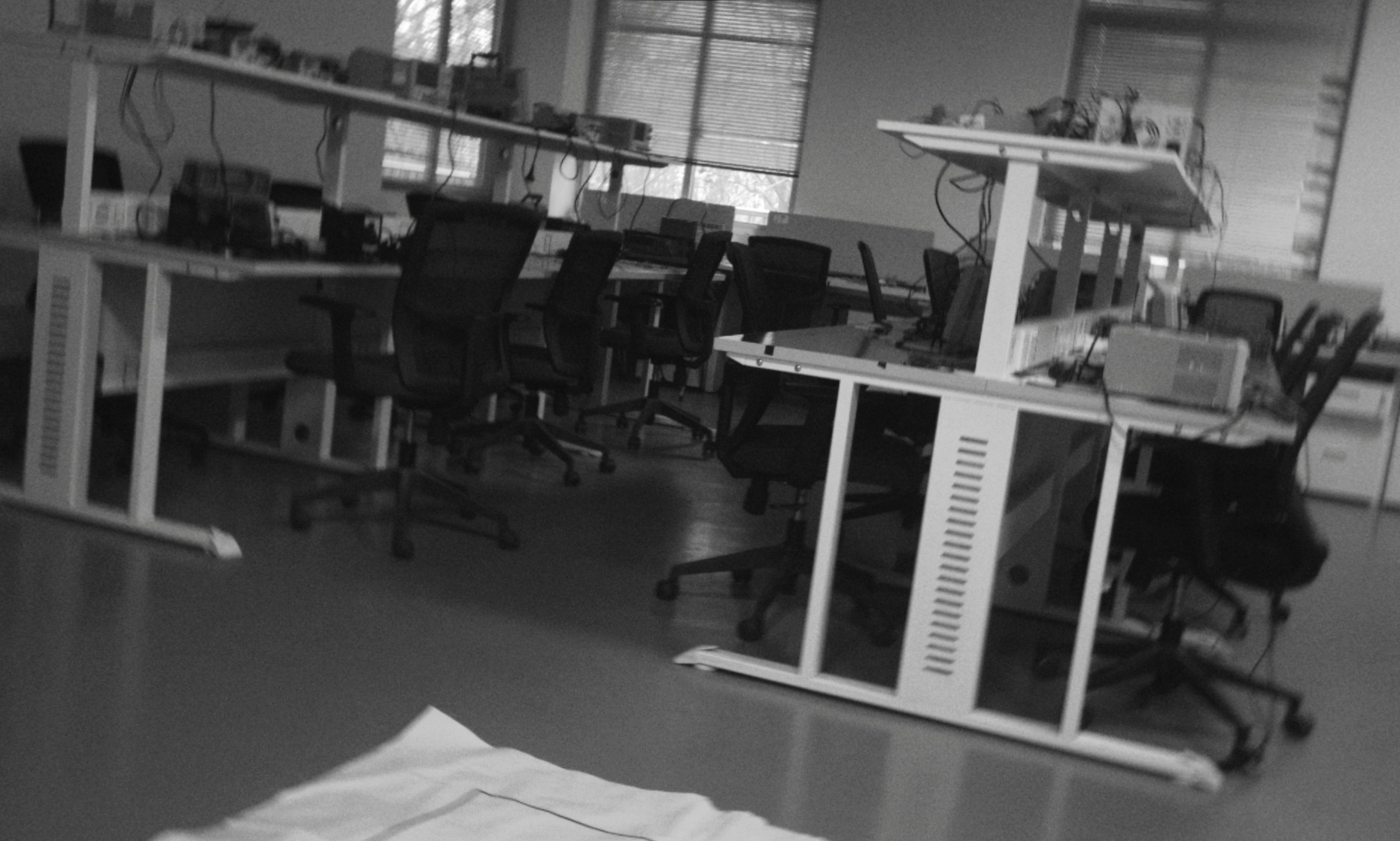}
    \caption{lab}
    \label{fig:lab}
  \end{subfigure}
  \begin{subfigure}[t]{0.32\linewidth}
      \setlength{\abovecaptionskip}{2pt}
  \includegraphics[width=\linewidth]{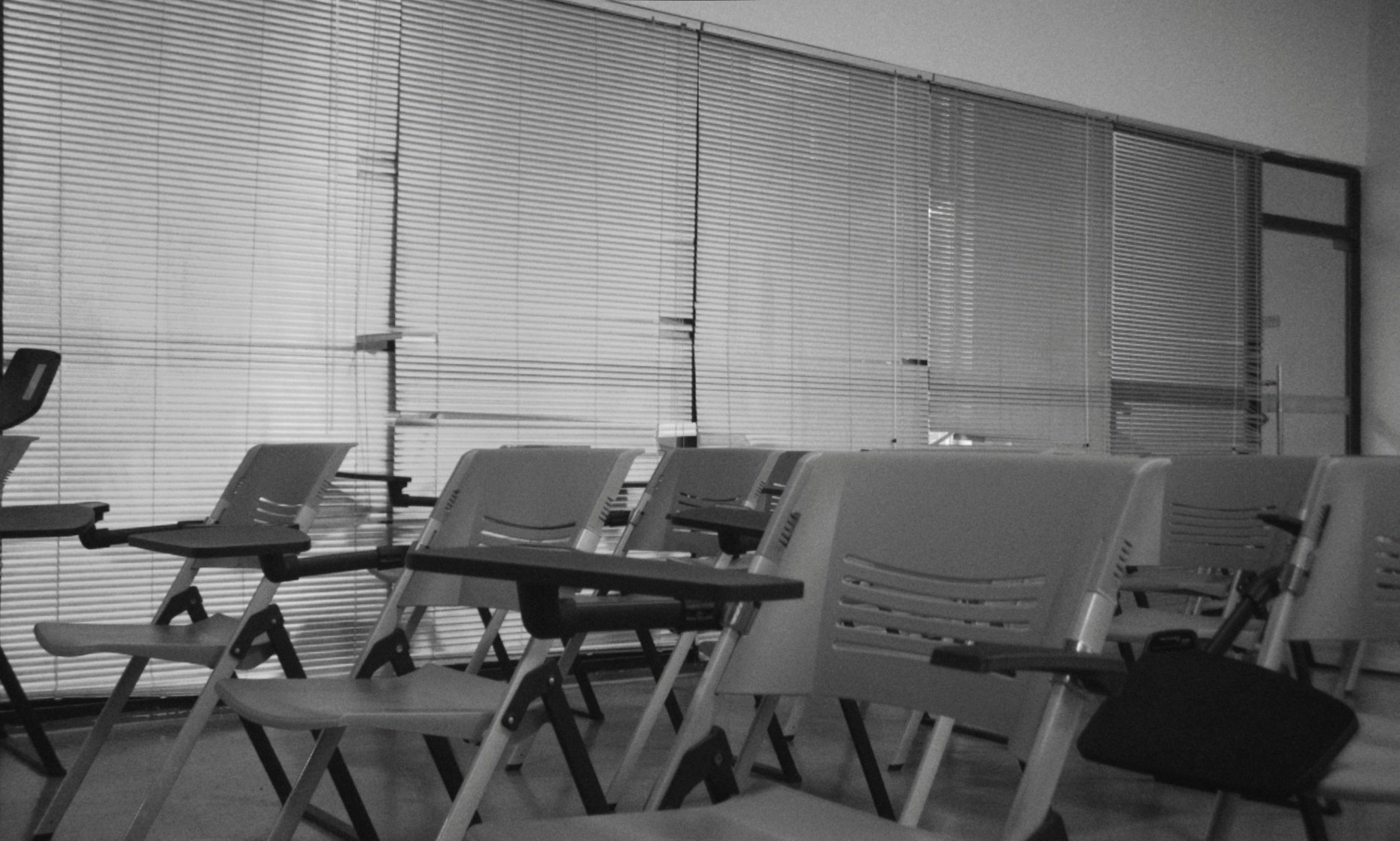}
  \caption{classroom}
  \label{fig:classroom}
  \end{subfigure}
  \caption{\mai{Self-collected data scenarios used for evaluation, including an AprilTag calibration board (serving as input for target-based methods in comparative experiments), as well as diverse environments such as a logo, boardroom, office, lab, and classroom (utilized as input for both our proposed method and the pure CCA method). }}
  \label{fig:self_data_scenarios}
  \vspace{-0.5mm}
\end{figure}

\section{Experiments} \label{sec:experiments}

\subsection{Datasets and Evaluation Metrics}
\label{subsec:datasets and evaluation metrics}
\textbf{ECD Dataset~\cite{Mueggler17ijrr}.}
The ECD dataset~\cite{Mueggler17ijrr} includes data sequences featuring handheld rotational motion across various scenes, which are recorded with the DAVIS~\cite{Brandli14ssc} sensor.
The DAVIS sensor integrates a 240 × 180 pixel event camera along with an IMU (InvenSense MPU-6150) which operates at a 1000 Hz output frequency.
In addition to event data, the DAVIS sensor is capable of outputting standard frames.
All the outputs, including the event stream, standard frames and the IMU measurements are synchronized at the hardware level.
Sequences \texttt{shapes\_rotation}, \texttt{poster\_rotation}, \texttt{boxes\_rotation}, and \texttt{dynamic\_rotation} are used to validate the effectiveness of our method, while the sequence \texttt{calibration} is used in the comparison with other methods.

\textbf{Self-collected Dataset.}
As shown in Fig.~\ref{fig:sensor_setup}, we build a multi-sensor platform that integrates the DAVIS346 event camera with a resolution of 346 × 260, the FLIR Blackfly S standard RGB camera with a resolution of 1920 × 1200 running at 20 Hz, the MTi-630 IMU that runs at 400 Hz, and the Livox Mid-360 LiDAR that runs at 10 Hz. 
In addition, we use the hardware-level clock synchronization scheme from~\cite{sun2024evttc_dataset}, and we employ the SyncOut signal from the MTi-630 IMU to provide a unified trigger for the event camera and standard RGB camera. Fig.~\ref{fig:self_data_scenarios} shows groups of data collected from six different scenarios, labeled as \texttt{AprilTag}, \texttt{logo}, \texttt{boardroom}, \texttt{office}, \texttt{lab}, and \texttt{classroom}, respectively. The aforementioned sequences each last approximately 150-second, \todo{with \texttt{AprilTag} being 6-DoF and the others being 3-DoF.}

\textbf{Evaluation Metrics.}
In this section, we will compare the results of rotation extrinsic parameters and time offset estimation from different methods. 
To make the comparison easier, the rotations are represented using rotation vectors.

\subsection{Evaluation Result on the Public Dataset}
\label{subsec:public evaluation results}
To demonstrate the effectiveness of our method, we compare it against alternative approaches that can estimate rotation extrinsic parameters and time offsets.
Kalibr~\cite{Furgale13iros} is a widely recognized camera-IMU calibration toolbox that utilizes a target-based method to estimate the extrinsic parameters and time offsets between cameras and IMUs. 
By using a known calibration target, Kalibr can accurately estimate the relative position and orientation between the camera and IMU, as well as the offset between their timestamps.
e2calib~\cite{Muglikar2021CVPR} is a learning-based method that first reconstructs images from event data and then uses the reconstructed images for the subsequent calibration process.
\begin{table}[!t]
\huge
\centering
\caption{Estimated rotation extrinsic parameters and time offsets between the event camera and the IMU on the ECD dataset.}
\renewcommand{\arraystretch}{1.3}
\resizebox{\columnwidth}{!}{
\setlength{\tabcolsep}{0.55em}
\begin{tabular}{ccccc}
\toprule
Seq. Name \mai{(method)} & X-axis (\textdegree) & Y-axis (\textdegree) & Z-axis (\textdegree) & T\_offset (ms)  \\
\midrule
shapes\_rotation \mai{(ours)} & 0.10& -1.01&  -1.07& -3.25\\
poster\_rotation \mai{(ours)} & -0.09& -1.15 & -1.06& -2.31\\
boxes\_rotation \mai{(ours)} & 0.54& -0.12& -1.10& -2.42\\
dynamic\_rotation \mai{(ours)} & -0.73& -0.54& -1.26& -2.84\\
\mai{calibration (Kalibr~\cite{Furgale13iros})} & 0.53 & -0.37 & -0.70 & -2.40 \\
\mai{calibration (e2calib~\cite{Muglikar2021CVPR})} & 0.32 & -1.13 & -0.73 & -1.10 \\


\bottomrule
\end{tabular}
}
\label{tab: ecd event imu res}

\begin{flushleft}
\vspace{-2mm}
\footnotesize \hspace{2mm}\todo{Calibration errors are not reported due to unavailability of ground truth.}
\end{flushleft}

\end{table}
\begin{table}[!t]
\huge
\centering
\caption{Estimated rotation extrinsic parameters and time offsets between the event data and the standard frame on the ECD dataset.}
\renewcommand{\arraystretch}{1.3}
\resizebox{\columnwidth}{!}{
\setlength{\tabcolsep}{0.55em}
\begin{tabular}{ccccc}
\toprule
Seq. Name \mai{(method)} & X-axis (\textdegree) & Y-axis (\textdegree) & Z-axis (\textdegree) & T\_offset (ms)  \\
\midrule
shapes\_rotation \mai{(ours)} & 0.87& -0.31& -0.40& -0.53\\
poster\_rotation \mai{(ours)} & 1.02& -0.34& 0.20& -0.40\\
boxes\_rotation \mai{(ours)} & 1.10& -0.13& -0.09& 0.34\\
dynamic\_rotation \mai{(ours)} & 0.08& -0.38& -0.55& -1.73\\
\mai{calibration (e2calib~\cite{Muglikar2021CVPR})} & -0.16 & 0.21 & 0.01 & 1.99 \\

\bottomrule
\end{tabular}
}
\label{tab: ecd event rgb res}
\vspace{-7mm}

\end{table}

Tab.~\ref{tab: ecd event imu res} and~\ref{tab: ecd event rgb res} present the event-IMU and event-frame calibration results across different sequences.
Since both Kalibr and e2calib are target-based methods, their results are obtained from the sequence \texttt{calibration}.
Given that event data and standard frames from the DAVIS sensor are captured using the same set of photodiodes, the calibration results from Kalibr on standard frames will serve as the reference for event-IMU calibration.
For this reason, the extrinsic rotation and temporal offsets between event data and standard frames should theoretically be zero.
For the results in the four rotation sequences, we use the first 30 seconds for the computation.
In the event-IMU calibration, our results are very close to Kalibr's estimation, with a difference of approximately 1 degree in the rotational extrinsic parameters and a maximum difference of around 1 ms in the time offset.
The images reconstructed by e2calib still exhibit differences compared to the standard frames, leading to inconsistencies between the calibration results of e2calib and Kalibr.
In the event-frame calibration, our estimated extrinsic rotation is close to zero, and the time offset estimation shows smaller errors compared to e2calib.
Note that in the dynamic\_rotation sequence, both the event-IMU and event-frame calibrations exhibit larger errors, likely caused by the presence of dynamic objects in this sequence, which affects the visual sensor's angular velocity estimation.
\subsection{Evaluation Result on Self-collected Data}
\label{subsec:self evaluation results}

\begin{table}[t]
\huge
\centering
\caption{Estimated rotation extrinsic parameters between the event camera and the IMU on the self-collected data.}
\renewcommand{\arraystretch}{1.3}
\resizebox{\columnwidth}{!}{
\setlength{\tabcolsep}{0.55em}
\begin{tabular}{ccccccccc}
\toprule
\multirow{2}{*}{Seq. Name} & \multirow{2}{*}{Method} & \multicolumn{2}{c}{X-axis (\textdegree)} & \multicolumn{2}{c}{Y-axis (\textdegree)} & \multicolumn{2}{c}{Z-axis (\textdegree)} \\
\cmidrule(l{1mm}r{1mm}){3-4} 
\cmidrule(l{1mm}r{1mm}){5-6} 
\cmidrule(l{1mm}r{1mm}){7-8} 
{~}& {~}& mean & std & mean & std & mean & std  \\
\midrule
\multirow{2}{*}{logo} & ours & 1.07& \textbf{0.08}&  -88.85& \textbf{0.17} & 0.82 & \textbf{0.12} \\
{~} & CCA~\cite{qiu2021tro} & 1.23 & 0.10&  -88.66 & 0.37 & 0.82 & 0.17 \\
\midrule
\multirow{2}{*}{boardroom} & ours & 0.12 & \textbf{0.29}&  -88.13& \textbf{0.23} & 1.87& \textbf{0.12} \\
{~} & CCA~\cite{qiu2021tro} & 0.10 & 0.42&  -87.84 & 0.59 & 1.92 & 0.35 \\
\midrule
\multirow{2}{*}{office} & ours & 1.83 & \textbf{0.24}&  -88.49& \textbf{0.48} & 1.38 & \textbf{0.14} \\
{~} & CCA~\cite{qiu2021tro} & 1.96 & 0.27&  -87.29 & 0.51 & 1.40 & 0.35 \\
\midrule
\multirow{2}{*}{lab} & ours & 0.73 & 0.48&  -89.35 & \textbf{0.21} & 1.42 & \textbf{0.16} \\
{~} & CCA~\cite{qiu2021tro} & 0.54 & \textbf{0.38}&  -89.20 & 0.52 & 1.36 & 0.39 \\
\midrule
\multirow{2}{*}{classroom} & ours & 0.37& \textbf{0.27}& -88.04 & \textbf{0.20}&  0.93 & 0.48 \\
{~} & CCA~\cite{qiu2021tro} & 0.10 & 0.42&  -87.84 & 0.59 & 1.92 & \textbf{0.35} \\
\midrule
\multirow{2}{*}{AprilTag} & Kalibr~\cite{Furgale13iros} & 1.10 & - & -88.33 & - & 0.99 & - \\
{~} & e2calib~\cite{Muglikar2021CVPR} & 1.12 & - & -88.34 & - & 0.96 & -\\

\bottomrule
\end{tabular}
}
\label{tab: self_data imu table}
\begin{flushleft}
\vspace{-2mm}
\footnotesize \hspace{2mm} \todo{Calibration errors are not reported due to unavailability of ground truth;}

\footnotesize \hspace{2mm} "-" represents not applicable.
\end{flushleft}

\vspace{-19mm}
\end{table}

\begin{table}[t]
\huge
\centering
\caption{Estimated rotation extrinsic parameters between the event camera and the standard frame on the self-collected data.}
\renewcommand{\arraystretch}{1.3}
\resizebox{\columnwidth}{!}{
\setlength{\tabcolsep}{0.55em}
\begin{tabular}{ccccccccc}
\toprule
\multirow{2}{*}{Seq. Name} & \multirow{2}{*}{Method} & \multicolumn{2}{c}{X-axis (\textdegree)} & \multicolumn{2}{c}{Y-axis (\textdegree)} & \multicolumn{2}{c}{Z-axis (\textdegree)} \\
\cmidrule(l{1mm}r{1mm}){3-4} 
\cmidrule(l{1mm}r{1mm}){5-6} 
\cmidrule(l{1mm}r{1mm}){7-8} 
{~}& {~}& mean & std & mean & std & mean & std  \\
\midrule
\multirow{2}{*}{logo} & ours & 0.79& \textbf{0.19}&  0.92& \textbf{0.23} & 0.07 & \textbf{0.09} \\
{~} & CCA~\cite{qiu2021tro} & 1.60 & 0.32&  0.76 & 0.57 & 0.13 & 0.25 \\
\midrule
\multirow{2}{*}{boardroom} & ours & -0.11 & \textbf{0.19} & 1.19 & \textbf{0.22} & -0.04& \textbf{0.12} \\
{~} & CCA~\cite{qiu2021tro} & 0.22 & 0.31&  1.64 & 0.57 & 0.17 & 0.30 \\
\midrule
\multirow{2}{*}{office} & ours & 1.23 & \textbf{0.44} &  1.68 & \textbf{0.40} & 0.54 & \textbf{0.19} \\
{~} & CCA~\cite{qiu2021tro} & 0.41 & 2.99&  5.05 & 6.73 & 0.52 & 0.56 \\
\midrule
\multirow{2}{*}{lab} & ours & 0.75 & \textbf{0.40}&  -0.89 & \textbf{0.19} & 0.24 & 0.29 \\
{~} & CCA~\cite{qiu2021tro} & 0.67 & 0.70&  -0.37 & 0.86 & 0.31 & \textbf{0.19} \\
\midrule
\multirow{2}{*}{classroom} & ours & 0.67& 0.36& 1.47 & \textbf{0.29}&  -0.35 & \textbf{0.29} \\
{~} & CCA~\cite{qiu2021tro} & 0.21 & \textbf{0.31}&  1.64 & 0.48 & 0.28 & 0.33 \\
\midrule
AprilTag & e2calib~\cite{Muglikar2021CVPR} & 0.29 & - & 0.69 & - & 0.23 & -\\


\bottomrule
\end{tabular}
}
\label{tab: self_data rgb table}
\end{table}

\begin{table}[t]
\huge
\centering
\caption{Estimated rotation extrinsic parameters between the event camera and the LiDAR on the self-collected data.}
\renewcommand{\arraystretch}{1.3}
\resizebox{\columnwidth}{!}{
\setlength{\tabcolsep}{0.55em}
\begin{tabular}{ccccccccc}
\toprule
\multirow{2}{*}{Seq. Name} & \multirow{2}{*}{Method} & \multicolumn{2}{c}{X-axis (\textdegree)} & \multicolumn{2}{c}{Y-axis (\textdegree)} & \multicolumn{2}{c}{Z-axis (\textdegree)} \\
\cmidrule(l{1mm}r{1mm}){3-4} 
\cmidrule(l{1mm}r{1mm}){5-6} 
\cmidrule(l{1mm}r{1mm}){7-8} 
{~}& {~}& mean & std & mean & std & mean & std  \\
\midrule
\multirow{2}{*}{logo} & ours & 70.78& \textbf{0.36}&  -69.38& \textbf{0.35} & 69.19 & \textbf{0.23} \\
{~} & CCA~\cite{qiu2021tro} & 70.37 & 1.17&  -69.82 & 0.79 & 70.60 & 0.90 \\
\midrule
\multirow{2}{*}{boardroom} & ours & 69.45 & \textbf{0.21}&  -67.98& \textbf{0.36} & 69.42& \textbf{0.22} \\
{~} & CCA~\cite{qiu2021tro} & 70.77 & 1.29&  -68.07 & 1.18 & 69.76 & 0.85 \\
\midrule
\multirow{2}{*}{office} & ours & 71.14 & \textbf{0.55}&  -68.39& \textbf{0.64} & 68.70 & \textbf{0.62} \\
{~} & CCA~\cite{qiu2021tro} & 70.04 & 1.00&  -67.68 & 1.09 & 68.44 & 1.63 \\
\midrule
\multirow{2}{*}{lab} & ours & 69.33 & \textbf{0.43}&  -69.34 & \textbf{0.28} & 70.36 & \textbf{0.26} \\
{~} & CCA~\cite{qiu2021tro} & 69.68 & 0.53& -69.35 & 0.97 & 70.92 & 0.94 \\
\midrule
\multirow{2}{*}{classroom} & ours & 69.86& \textbf{0.63}& -68.63 & \textbf{0.80}&  68.53 & 1.00 \\
{~} & CCA~\cite{qiu2021tro} & 70.77 & 1.29&  -68.07 & 1.18 & 69.76 & \textbf{0.85} \\


\bottomrule
\end{tabular}
}
\label{tab: self_data lidar table}
\vspace{-2mm}
\end{table}

\begin{table}[t]
\huge
\centering
\caption{Estimated time offsets between the event camera and the IMU,  standard camera, and LiDAR on the self-collected data.}
\renewcommand{\arraystretch}{1.3}
\resizebox{\columnwidth}{!}{
\setlength{\tabcolsep}{0.55em}
\begin{tabular}{ccccccccc}
\toprule
\multirow{2}{*}{Seq. Name} & \multirow{2}{*}{Method} & \multicolumn{2}{c}{IMU (ms)} & \multicolumn{2}{c}{Camera (ms)} & \multicolumn{2}{c}{LiDAR (ms)} \\
\cmidrule(l{1mm}r{1mm}){3-4} 
\cmidrule(l{1mm}r{1mm}){5-6} 
\cmidrule(l{1mm}r{1mm}){7-8} 
{~}& {~}& mean & std & mean & std & mean & std  \\
\midrule
\multirow{2}{*}{logo} & ours & 1.81 & \textbf{0.29} &  0.34 & 1.17 & 4.27& \textbf{2.11} \\
{~} & CCA~\cite{qiu2021tro} & 1.22 & 0.38 &  -0.16 & \textbf{0.73} & 4.53 & 3.04 \\
\midrule
\multirow{2}{*}{boardroom} & ours & 2.27 & \textbf{0.49}&  0.32 & \textbf{1.79} & 4.53 & \textbf{1.52} \\
{~} & CCA~\cite{qiu2021tro} & 0.83 & 0.85 &  0.30 & 2.00 & -1.77 & 3.70 \\
\midrule
\multirow{2}{*}{office} & ours & 1.68& \textbf{0.48} &  -0.52 & \textbf{0.95} & 2.57& 2.74 \\
{~} & CCA~\cite{qiu2021tro} & 1.93 & 0.59 &  -2.87 & 2.30 & 4.90 & \textbf{2.15} \\
\midrule
\multirow{2}{*}{lab} & ours & 1.27 & \textbf{0.58}&  -1.06 & \textbf{0.93} & 2.25& \textbf{1.05} \\
{~} & CCA~\cite{qiu2021tro} & 0.69 & 1.81&  -0.99 & 1.08 & -1.56 & 3.77 \\
\midrule
\multirow{2}{*}{classroom} & ours & 2.00 & \textbf{0.46}& -0.91 & \textbf{0.86}&  4.08& \textbf{1.42} \\
{~} & CCA~\cite{qiu2021tro} & 0.83 & 0.83&  -0.33 & 1.63 & -1.78 & 3.70 \\
\midrule
\multirow{2}{*}{AprilTag} & Kalibr~\cite{Furgale13iros} & 0.19 & - & - & - & - & - \\
{~} & e2calib~\cite{Muglikar2021CVPR} & 3.30 & - & 0.23 & - & - & -\\

\bottomrule
\end{tabular}
}
\label{tab: self_data time offset table}
\end{table}

\todo{In this part, we conduct the same comparative experiments outlined in Section~\ref{subsec:public evaluation results}, using the self-collected sequence~\texttt{AprilTag} as input for both Kalibr and e2calib, and the other self-collected sequences for our method.}
\todo{We randomly extract 30 second segments from each sequence as input for our method, repeating this process 100 times.}
\todo{The extrinsic rotation results for event-IMU, event-frame, and event-LiDAR on self-collected data are shown in Tab.~\ref{tab: self_data imu table},~\ref{tab: self_data rgb table} and~\ref{tab: self_data lidar table}, and the time offset results are shown in Tab.~\ref{tab: self_data time offset table}. }
\todo{In the extrinsic rotation calibration of event-IMU and event-frame, the extrinsic parameters are very close to those obtained using Kalibr, with differences of no more than 1 degree.
Similarly, the time offsets of event-IMU and event-frame are closely aligned with those of e2calib, consistent with the results on the public dataset. 
Both the extrinsic rotation and time offset demonstrate smaller variance and greater stability across all three setups compared to those obtained with the pure CCA method.
However, regardless of whether using our method or the pure CCA method, the result from event-LiDAR is less stable than those from the other two setups.}
This instability may be caused by motion distortion in the point cloud~\cite{qiu2021tro}. 
Although using motion compensation based on the constant velocity model can partially address the distortion, the model fails when the speed is too high. 
Therefore, motion compensation for LiDAR point clouds is an optional item in our code.
\begin{figure}[!t]
  \centering
  \begin{subfigure}[t]{0.49\linewidth}
    \setlength{\abovecaptionskip}{1pt}
    \includegraphics[width=\linewidth]{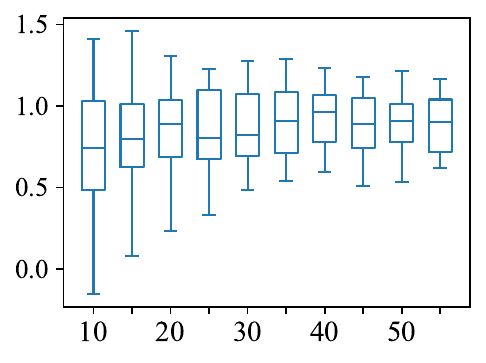}
    \caption{X-axis}
  \end{subfigure}
  \begin{subfigure}[t]{0.49\linewidth}
      \setlength{\abovecaptionskip}{1pt}

    \includegraphics[width=\linewidth]{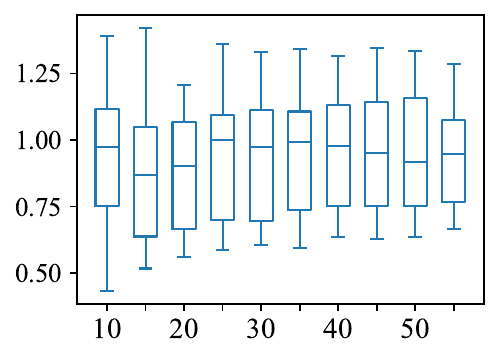}
    \caption{Y-axis}
  \end{subfigure}
  \begin{subfigure}[t]{0.49\linewidth}
      \setlength{\abovecaptionskip}{1pt}

    \includegraphics[width=\linewidth]{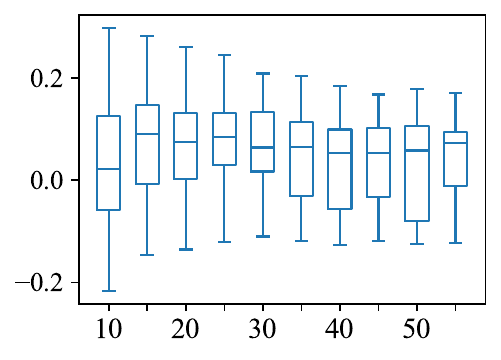}
    \caption{Z-axis}
  \end{subfigure}
  \begin{subfigure}[t]{0.49\linewidth}
  \setlength{\abovecaptionskip}{1pt}
  \includegraphics[width=\linewidth]{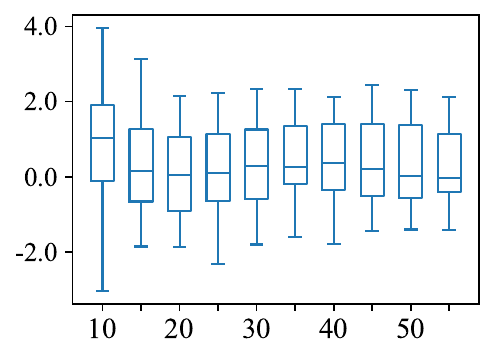}
  \caption{Time Offset}
  \end{subfigure}
  \caption{The influence of different sequence durations on our approach. The x-axis represents the sequence duration (s), the y-axis of (a), (b) and (c) represent the extrinsic rotation (\textdegree), and the y-axis of (d) represents the time offset (ms).}
  \label{fig: diff duration compare}
\end{figure}

\todo{In addition, by testing the event and IMU data in \texttt{logo},} Fig.~\ref{fig: diff duration compare} illustrates the influence of different sequence durations on our approach. 
As the sequence duration increases, the results tend to stabilize.
The longer the data duration, the less the impact of outliers in the speed estimation results on calibration, \mai{and the processing time is extended.} 
Therefore, adopting a 30-second interval in our method is comparatively reasonable. 
In conclusion, our method achieves similar levels of accuracy as these target-based approaches \todo{while demonstrating superior stability over pure CCA-based methods}.



\section{conclusion} 
\label{sec:conclusion}

This paper provides a target-free solution to the problem of temporal and rotational extrinsic calibration for event-centric multi-sensor systems.
To get rid of event-to-image conversion, the proposed motion-based method leverages the kinematic correlation across rotational motion 
estimates obtained from heterogeneous sensors, respectively.
The entire pipeline consists of a CCA-based initialization followed with a joint nonlinear refinement using a continuous-time parametrization in SO(3).
Extensive validation on real-world data demonstrates that our method not only achieves calibration accuracy on par with target-based approaches but also exhibits greater stability compared to purely CCA-based methods. 
This results in enhanced flexibility, stability, and robustness, positioning our approach as a tailored solution for event-centric configuration.
We wish the resulting calibration toolbox will serve as a standard tool in the community and facilitates robotic applications that involve event-based cameras.

\bibliographystyle{IEEEtran} 
\bibliography{myBib}

\end{document}